\documentclass{article}
\usepackage{spconf,amsmath,graphicx}

\usepackage{amsfonts}
\usepackage{url}
\usepackage{multicol}
\usepackage{multirow}
\usepackage{hyperref}
\usepackage{cleveref}
\usepackage[font=small,skip=0pt]{caption}
\usepackage{subfig}
\usepackage{xcolor}
\usepackage{amsthm}         
\usepackage{xfrac}

\title{Correcting Faulty Road Maps by Image Inpainting}

\name{
    Soojung Hong\textsuperscript{1}$^{\dag*}$\thanks{$^\dag$Work done at HERE Technologies. $^*$Equal contributors.}
    Kwanghee Choi\textsuperscript{2}$^*$    
}

\address{
    ETH Z{\"u}rich, 
	Dept. of Computer Science, Z{\"u}rich, Switzerland\textsuperscript{1} \\
    Carnegie Mellon University, Language Technologies Institute, USA\textsuperscript{2}
}

\begin{document}
\newcommand{\fix}[1]{\textcolor{red}{#1}}

\maketitle


\begin{abstract}
As maintaining road networks is labor-intensive, many automatic road extraction approaches have been introduced to solve this real-world problem, fueled by the abundance of large-scale high-resolution satellite imagery and advances in computer vision.
However, their performance is limited for fully automating the road map extraction in real-world services.
Hence, many services employ the two-step human-in-the-loop system to post-process the extracted road maps: error localization and automatic mending for faulty road maps.
Our paper exclusively focuses on the latter step, introducing a novel image inpainting approach for fixing road maps with complex road geometries without custom-made heuristics, yielding a method that is readily applicable to any road geometry extraction model.
We demonstrate the effectiveness of our method on various real-world road geometries, such as straight and curvy roads, T-junctions, and intersections.
\end{abstract}


\section{Introduction} \label{sec:intro}

Developing and maintaining accurate road maps is essential for many location-based applications.
However, keeping the road maps up-to-date is an arduous process that often requires extensive manual labor \cite{miller2014huge}.
Hence, recent advances have been dominated by utilizing aerial and satellite imagery that offer abundant, reliable, high-resolution, and real-time data \cite{zhang2017road,liu2022survey}.
Its properties are ideal for deep learning-based approaches such as semantic segmentation \cite{garcia2017review}, which predicts the existence of roads.
For example, Cira et al. \cite{ijgi11010043} leverages image inpainting techniques by training the segmentation model on the synthetic noise.
Others leverage the GPS (Global Positioning System) trace points to extract road shapes \cite{ijgi6120403,ijgi6120404}.

To further enhance the performance of the road map extraction models, post-processing methods are often applied \cite{ijgi11010043,sujatha2015connected,wang2016new,bakhtiari2017semi}.
While delegating the extraction of the overall geometry of roads to the upstream models, post-processing methods concentrate on the finer details: restoring the missing connections between road segments, removing noisy predictions, and generating final binary maps from real-valued predictions.
However, in many cases, heuristics are often too simple to tackle real-world road structures with complex geometry \cite{6738568}, resulting in performance degradation \cite{SOTA_article}.
Furthermore, non-road objects such as vegetation, occlusions, and shadows \cite{mnih2010learning,manandhar2019towards} make reliable extraction challenging.

\begin{figure}[t!]
  \centering
  \subfloat[Straight Road]{\includegraphics[width=0.24\textwidth]{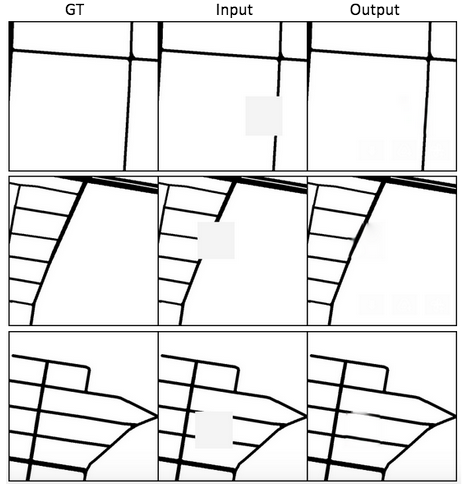}}
  \subfloat[Curvy Road]{\includegraphics[width=0.24\textwidth]{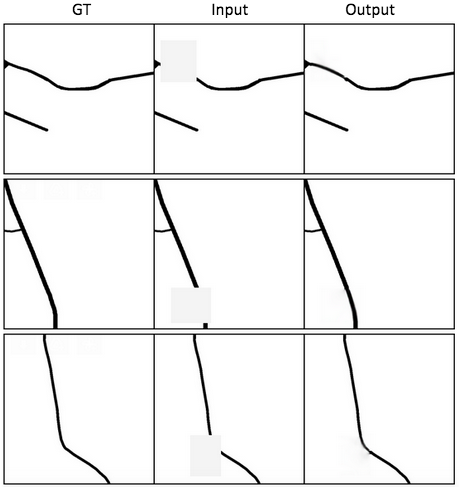}}\\
  \vspace{-1em}
  \subfloat[T-junction]{\includegraphics[width=0.24\textwidth]{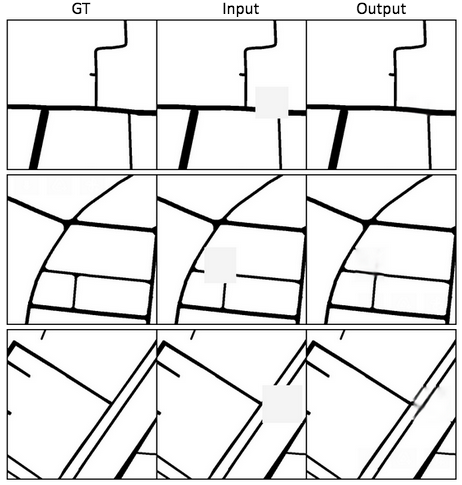}}
  \subfloat[Intersection]{\includegraphics[width=0.24\textwidth]{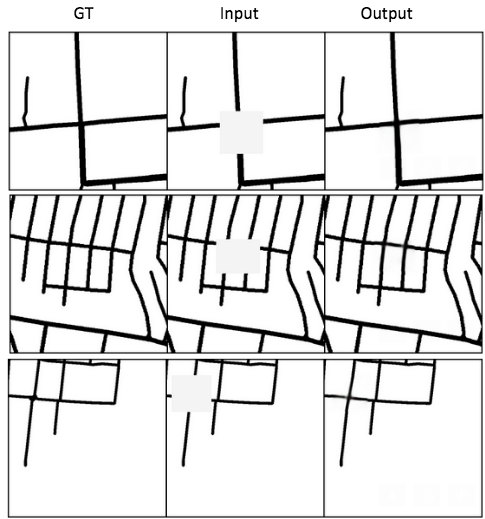}}
  \caption{
  GT, Input, and Output denotes the ground truth, the input image that our model receives, and its road inpainting result.
  Our method successfully reconstructs the real-world road types including T-junctions and intersections by capturing the geometric information of the surroundings.
  }
  \label{fig:qualitative_analysis}
\end{figure}

To tackle the limitations above, human-in-the-loop approaches are often used in production-level applications, making the manual labor of labeling road maps much more efficient via automation.
A two-step approach is commonly employed to maintain existing road maps: error localization and automatic mending \cite{ijgi10010008}.
Error localization recommends locations to be fixed to the human annotators, where automatic mending tries to correct the road maps, giving a good prior for the human annotators to fix further.
By dividing post-processing into two consecutive steps, human annotators can easily intervene within the pipeline, which is ideal for the human-in-the-loop system.

For the automatic mending step, one often employs heuristics that are tailored toward each specific upstream model, such as morphological operation techniques \cite{sujatha2015connected,wang2016new,bakhtiari2017semi} or polynomial curve fitting \cite{Liu2019MultiscaleRC}.
However, heuristics cannot improve the quality from a road-geometric perspective, as it cannot fix the predictions if the provided road geometry is wrong.
Further, dependency on the upstream segmentation model increases the complexity of the whole system.

Our paper focuses on improving the automatic mending step by introducing a novel data-driven method that leverages the abundant amount of satellite imagery available.
Unlike the hand-crafted custom heuristics, our method can continuously improve as road map data accumulates while removing the dependency on the road map extraction models, decreasing the overall system complexity.
We borrow the image inpainting techniques \cite{ijgi11010043,ORG_4,ORG_13} by regarding a faulty map region to be an area to be inpainted.
Our method is intuitive and easily applicable to existing systems, not requiring bells and whistles.
To demonstrate the effectiveness of our approach, we separately evaluate complex structures such as curvy roads, T-junctions, and intersections, where existing heuristic methods are known to fail \cite{6738568}.
Our experiments qualitatively and quantitatively confirm our approach's effectiveness.

To summarize, our contributions are: (1) introduce a data-driven method independent of the upstream road map extraction models to mend faulty road maps based on image inpainting, and (2) demonstrate the effectiveness of our approach via extensive evaluation on various road geometries.

\section{Automatically Correcting Faulty Road Map Predictions} \label{sec:method}
We introduce a novel image inpainting approach for mending faulty road map predictions motivated by the following idea: \textit{Given the abundance of existing images of road maps, why not teach the geometric specifics of roads to a neural network in a self-supervised manner?}
Based on this intuition, we leverage the existing image inpainting model and automatically apply modifications to serve the geometric characteristics of road maps.
We first start with the pilot study to motivate our design decisions in \cref{subsec:ablation}, where we describe the model architecture in \cref{subsec:model_arch} and the customized loss in \cref{subsec:method_loss}.

\subsection{Challenges of inpainting road maps} \label{subsec:ablation}
As a pilot study, we utilized the state-of-the-art diffusion-based image inpainting model, RePaint \cite{lugmayr2022repaint}, for the road maps.
We first trained the underlying Denoising Diffusion Probabilistic Model (DDPM) \cite{ho2020denoising} with the road map dataset (details in \cref{subsec:dataset}) for 300k iterations with the default settings of DDPM and RePaint.
By using the trained DDPM model as a generative prior, RePaint samples pixels from the unmasked regions within the given road map to fill in the masked region.


\begin{figure}[t]
  \centering
  \includegraphics[width=0.48\textwidth]{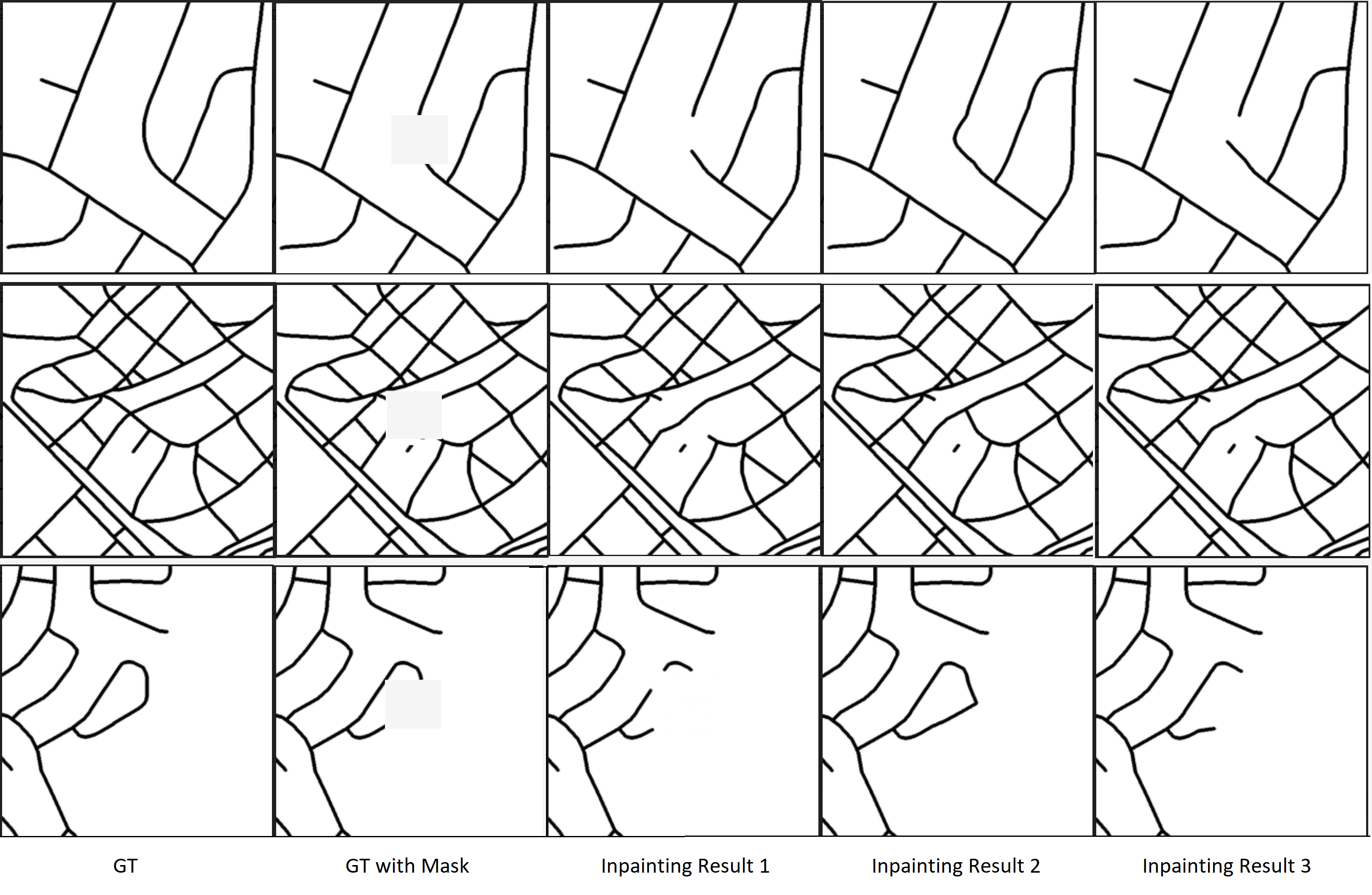}
  \caption{
Road geometry inpainting using RePaint.
RePaint fails to yield geometrically sensible road maps and produces random road maps with different random seed.
}
  \label{fig:repaint_based_inpainting}
\end{figure}

\Cref{fig:repaint_based_inpainting} shows that, even though the inpainted road maps are crisp, RePaint injects randomness during the sampling step by design \cite{lugmayr2022repaint}.
This randomness complicates the usage of the model in our problem's production setting.
Further, inpainted road segments by RePaint often fail to harmonize the overall geometry, especially for complex road geometries.
Additionally, RePaint is computationally heavy \cite{lugmayr2022repaint}; inference takes a few minutes per sample on a single GPU.

By observing the RePaint results, we obtain design requirements for the road map inpainting approach.
First, road map images have different inherent characteristics compared to natural images, such as CelebA-HQ \cite{Celeb}, which are commonly used to evaluate generative models.
Inpainted segments have to be geometrically sensible; simply mimicking the road segment patterns is not enough.
Also, unlike applications for creative purposes, its inpainted results should be stable and quick during inference.
Based on the requirements, we design our method as the following subsections.

\subsection{Model Architecture} \label{subsec:model_arch}
Our model aims to inpaint the geometric road shape in the target area.
As mentioned in \cref{sec:intro}, we assume the erroneous location is already localized.
We modify the model architecture of the Globally and Locally Consistent Image Completion (GLCIC) \cite{ORG_4} to capture the spatial relationship between the target road segment and the surrounding geometries while enjoying fast inference time with few seconds per sample on a single GPU.
We denote our model architecture, Globally and Locally Consistent Road map Completion (GLCRC), which consists of a generator and a discriminator.

\subsubsection{Generator with Deep Dilated Convolutions}
The generator has 6 conv layers followed by 8 dilated conv layers, with an increasing dilation factor from 2 to 9.
Increasing factors enable the model to learn the spatial context via bigger receptive fields, capturing the geometric relationship between the road shape of the area to be inpainted (\textit{i.e.} target area) and the surroundings.
Unlike natural images, which often have the freedom of inaccuracy at a granular level, the road geometries must be meticulously accurate, harmonizing with both neighboring roads and the overall geometry.

\subsubsection{Multi-scale Context Discriminator}
The discriminator is based on the multi-scale discriminator architecture \cite{ORG_4}, containing a global and a local context discriminator. 
We utilize this architecture to handle the global and local context of the road maps, grasping the overall geometry with its details.
The global context discriminator takes 256 $\times$ 256 input and feed-forward to 6 conv layers and the final linear layer.
The local context discriminator takes 128 $\times$ 128 image patch, which is a target region to fill in.
We concatenate two context vectors to pass through a linear layer and a sigmoid activation to represent the probability that a given inpainted image is plausible or not.

\subsection{Loss function} \label{subsec:method_loss}
To achieve better geometric line quality, we applied perceptual loss \cite{ORG_6} to avoid blurriness and Relativistic Least Square GAN (RaLSGAN) Loss \cite{ORG_7} to achieve sharper maps.
The goal of loss functions is to achieve correctness while increasing the sharpness of the inpainted road geometries.

\subsubsection{Avoiding Blurriness via Perceptual Loss}
\begin{figure}[t]
  \centering
  \subfloat{\includegraphics[width=0.48
  \textwidth]{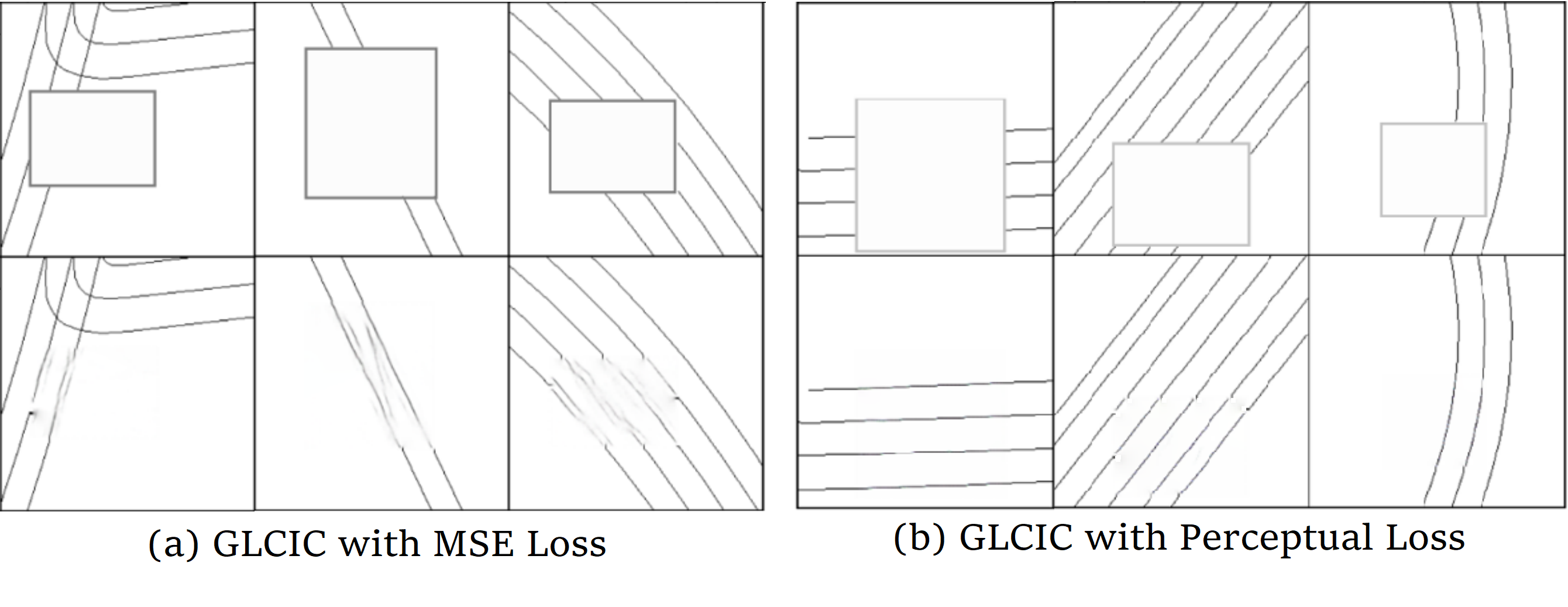}}
  \caption{
    Decreased road map blurriness by utilizing the perceptual loss: (a) GLCIC with MSE loss, and (b) GLCIC with perceptual loss.
    We mask the road maps (rectangles in the first row) so that the generator can reconstruct the roads within the mask (second row).
  }
  \label{fig:mse_is_blurry}
\end{figure}

The generator network is trained with the perceptual loss \cite{ORG_6} unlike the original GLCIC, which uses Mean Square Error (MSE) loss.
We observed that MSE loss is suboptimal for the road maps compared to the perceptual loss, as it produces more blurry road predictions (\Cref{fig:mse_is_blurry}), which further supports previous findings \cite{zhang2022soup}.
Note that the sharpness of the inpainted geometric lines depends on the degree of texture distortion along the road lines.
In addition to the sharpness of the inpainted geometric lines, the perceptual loss helped the model to generate the correct texture distortion along the road geometry, which makes realistic road shapes.

\subsubsection{Sharper Maps via Relativistic Least Square GAN (RaLSGAN) Loss}
For both global and local context discriminators, we adopt RaLSGAN loss \cite{ORG_7}, unlike the vanilla GLCIC model, which uses the Binary Cross Entropy loss.
Empirical findings indicate that RaLSGAN loss helps generate sharper lines \cite{ORG_7}, a desirable characteristic for road maps.

\section{Experiments} \label{sec:experiments}
We experimentally demonstrate the effectiveness of our approach by comparing three cases: (1) Vanilla GLCIC, (2) GLCRC (architectural modifications on GLCIC, \cref{subsec:model_arch}), and (3) GLCRC with better losses (\cref{subsec:method_loss}).
We provide the source codes of GLCRC for additional details.\footnote{\url{https://github.com/SoojungHong/image_inpainting_model_for_lane_geomery_discovery}}
We construct the remaining section as follows.
We first describe the dataset and the training details (\cref{subsec:dataset}).
Then, we show the experimental results on various road types (\cref{subsec:result}).

\subsection{Dataset} \label{subsec:dataset}
All the models were trained and evaluated with Massachusetts Road Geometry data \cite{MnihThesis}, which covers over 2600 square kilometers with diverse rural, suburban, and urban areas \cite{SOTA_article}.
The data reflects the real-world road geometries with a 1m spatial resolution and 1500 $\times$ 1500 pixel size.
We divided each image into nine 500 $\times$ 500 images, then resized into 256 $\times$ 256.
In all the experiments, we trained and tested the model with 9972 and 567 images, respectively.

\begin{table}[t]
  \centering
  \resizebox{1.0\linewidth}{!}{%
  \begin{tabular}{@{}lccc@{}}
    \hline\hline 
    Method & Correctness & Completeness & Quality \\
    \hline
    Vanilla GLCIC & 0.787 & 0.803 & 0.664 \\
    GLCRC & 0.789 & 0.811 & 0.668 \\
    GLCRC+L (Ours) & \textbf{0.795} & \textbf{0.831} & \textbf{0.671} \\

    \hline
  \end{tabular}
    }
    \vspace{0.5em}
  \caption{
  Road map fixing performance.
  GLCRC modified the architecture of vanilla GLCIC, and GLCRC+L further utilizes perceptual and RaLSGAN losses.
  }
  \label{tab:1}
\end{table}

\begin{figure*}
  \centering
  \subfloat[Intersection]{\includegraphics[height=0.34\textwidth]{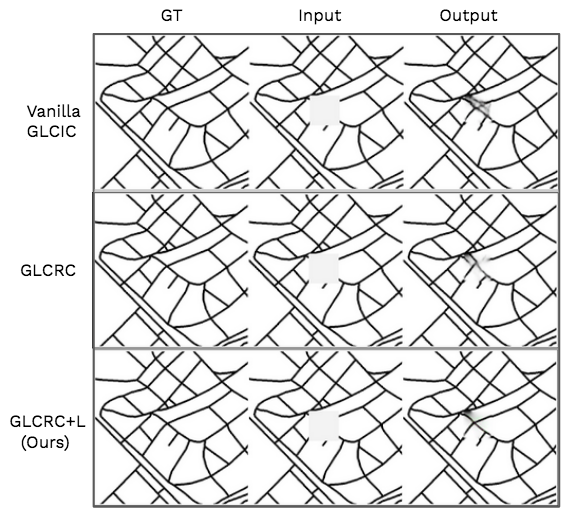}}
  \subfloat[Curvy road]{\includegraphics[height=0.34\textwidth]{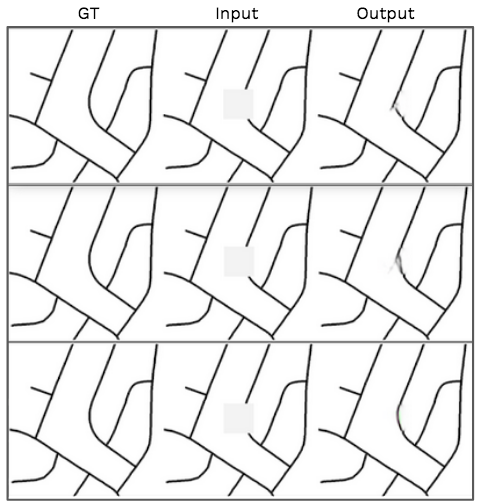}}
  \subfloat[Curvy road]{\includegraphics[height=0.34\textwidth]{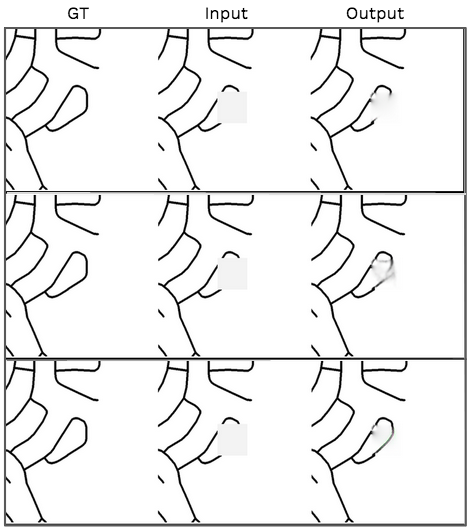}}
  \vspace{1em}
  \caption{
  Road inpainting results of Vanilla GLCIC (Baseline), GLCRC (Enhanced Architecture), and GLCRC+L (Ours, GLCRC with perceptual loss and RaLSGAN loss) on various road types, (a) Intersection, and (b, c) Curvy road.}
  \label{fig:comp_result}
\end{figure*}

\subsection{Training details} \label{subsec:training_details}
We followed the same training scheme as GLCIC \cite{ORG_4}.
We pretrained the generator and the discriminator network separately for 90k and 40k steps, respectively.
This pretraining approach is often used to train GANs \cite{ORG_4} to avoid the instability of adversarial training.
After completing the pretraining of each network, the generator and the discriminator are trained in an alternating manner for 90k iterations.

\begin{table}[t]
  \centering
  \resizebox{0.99\linewidth}{!}{%
  \begin{tabular}{@{}c cccc@{}}
    \hline\hline 
    Road type & Method & Correctness & Completeness & Quality \\
    \hline\
    \multirow{3}{*}{Straight} & Vanilla GLCIC & 0.787 & 0.786 & 0.649  \\
    & GLCRC & 0.750 & 0.806 &	0.635\\
    & GLCRC+L (Ours) & \textbf{0.894} & \textbf{0.898} & \textbf{0.811} \\
    \hline\
    \multirow{3}{*}{Curvy} & Vanilla GLCIC & \textbf{0.762} & 0.757 & \textbf{0.613} \\
    & GLCRC & 0.723 & \textbf{0.789} & 0.606 \\
    & GLCRC+L (Ours) & 0.754 & 0.766 & \textbf{0.613} \\
    \hline\
    \multirow{3}{*}{T-junction} &  Vanilla GLCIC & 0.775	& 0.788 & 0.642 \\
    & GLCRC & 0.785 & 0.792 &	0.651\\
    & GLCRC+L (Ours) & \textbf{0.842} & \textbf{0.849} & \textbf{0.733} \\
    \hline\
    \multirow{3}{*}{Intersection} & Vanilla GLCIC & 0.775	& 0.788 &	0.642\\
    & GLCRC & 0.785	& 0.792	& 0.651\\
    & GLCRC+L (Ours) & \textbf{0.786}& \textbf{0.793} &	\textbf{0.652} \\
    \hline\
  \end{tabular}
  }
  \caption{
  Comparing road map fixing performance on various road geometries.
  GLCRC denotes architectural modifications to the vanilla GLCIC and GLCRC+L denotes GLCRC with perceptual and RaLSGAN loss.
  }
  \label{tab:2}
\end{table}

\subsection{Experimental Results}\label{subsec:result}
We conduct quantitative (\Cref{tab:1}) and qualitative analysis (\Cref{fig:qualitative_analysis}) on whether the modifications are effective.
We use three metrics, Correctness, Completeness, and Quality, which are widely used in road extraction tasks \cite{wang2016new}.
Further, we also evaluate on various road types (\Cref{tab:2}) to verify the method's robustness.
We choose four random road maps per each road type: Straight, Curvy, T-junction, and Intersection, where the latter three types are known to be challenging for existing post-processing methods \cite{app12125873, rs13081417}.

\Cref{tab:1} shows the performance impact on various modifications: GLCRC being the architectural and GLCRC+L being the training loss modifications.
We can clearly observe that our approach shows the best image quality in the road geometry inpainting problem.
Furthermore, \Cref{fig:qualitative_analysis} shows the effectiveness of our method on various road types, implying that it understands underlying road geometries to reconstruct the road location.

We also compare the model and loss modifications on separate road types qualitatively (\Cref{fig:comp_result}) and quantitatively (\Cref{tab:2}).
\Cref{fig:comp_result} demonstrates that complex road shapes such as multiple T-junction areas and irregular curvy roads are successfully handled via our method.
Even though architectural changes can improve the overall road shapes, one also has to employ perceptual loss and RaLSGAN loss to enhance the smoothness and the sharpness of roads.
Further, \Cref{tab:2} indicates that ours shows superior performance in all the complex road types and evaluation metrics.

\section{Conclusion} \label{sec:conclusion}
Even though numerous advances have been made to the automated extraction of road maps from satellite imagery fueled by the abundance and availability of data, many still rely on hand-crafted heuristics and algorithms to fix faulty road map predictions.
Our work shed new light on the road-mending techniques motivated by the data-driven machine learning algorithms that prevail in vision applications today.
We introduce a novel method that regards faulty road map predictions as the image to inpaint.
We modify the existing image inpainting approaches to handle the geometric context while maintaining accuracy and crispness for road maps.
Our method does not require specific tuning on the numerous real-world road types nor the knowledge of the upstream road map extraction models; it only needs a sufficient amount of data to enhance the predictions.
Experiments demonstrate that our method is robust to various real-world road geometries such as curvy, T-junction, and intersection, which are challenging for existing post-processing methods.

\bibliographystyle{IEEEbib}
\bibliography{refs}

\end{document}